\documentclass{Interspeech}
\usepackage{url}

\interspeechcameraready

\title{Phonetic forced alignment for low-resource language varieties: \\Model training and evaluation on Chengdu Mandarin}

\author[affiliation={1}]{Zhiheng}{Qian}
\author[affiliation={2}]{Aini}{Li}
\author[affiliation={3}]{Hai}{Hu}
\author[affiliation={4}]{Liang}{Zhao}

\affiliation{}{Shanghai Jiao Tong University; $^2$City University of Hong Kong; \\$^3$The Hong Kong Polytechnic University; $^4$Beijing Foreign Studies University}{}
\email{n1vnhil@sjtu.edu.cn, ainili@cityu.edu.hk, hai.hu@polyu.edu.hk, liang.zhao@bfsu.edu.cn}

\usepackage{tabularx}
\usepackage{comment}
\usepackage{tipa}
\usepackage{array}        
\usepackage{enumitem}     
\usepackage{booktabs}
\usepackage{CJKutf8}
\usepackage{xcolor}
\usepackage{multirow}
\usepackage{hyperref}

\usepackage{arydshln}

\begin{document}

\maketitle


\begin{abstract}
Phonetic forced alignment is a key technique in phonetic research, yet existing alignment systems lack specialized models for low-resource language varieties. We address this by training text-dependent and text-independent aligners for Chengdu Mandarin using a 17-hour corpus and a custom G2P dictionary. We trained a text-dependent GMM-HMM model (Chengdu-MFA) and fine-tuned a pretrained audio encoder on frame classification with Chengdu-MFA's pseudo label for text-independent alignment (Chengdu-FC). Evaluation on an expert-annotated test set show that both methods significantly outperform Standard Mandarin baselines. Chengdu-MFA reduced average phone boundary differences by 31.8\%, while Chengdu-FC achieved a 61.2\% reduction. This work establishes a practical bootstrapping pipeline for developing accurate aligners for under-resourced varieties without labor- and time-intensive manual annotation.  

\end{abstract}

\section{Introduction}
The increasing availability of spoken language data has heightened the need for reliable automated methods in phonetic analysis. Phonetic forced alignment is a critical technique for synchronizing speech transcription with audio at the utterance, word, and phone levels \cite{yuan2023using, chodroff2025comparing}. By automating time-aligned annotations, dedicated aligners such as the Penn Forced Aligner \cite{yuan2008speaker}, the Prosodylab-Aligner \cite{gorman2011prosodylab}, FAVE \cite{rosenfelder2014fave}, the Montreal Forced Aligner (MFA) \cite{mcauliffe2017montreal}, and Charsiu Forced Aligner \cite{Zhu2022PhoneToAudio}, have greatly facilitated large-scale phonetic and sociolinguistic research. However, existing models for phonetic forced alignment are predominantly trained on standardized, high-resource languages (e.g., Standard Mandarin). When applied to regional or non-standard language varieties, their performance may decline due to systematic differences in the sound systems. Although some alignment toolkits allow researchers to train custom models, doing so from scratch is particularly challenging for low-resource varieties, which typically lack both massive speech corpora and specialized phonetic dictionaries.

This presents a significant methodological bottleneck: how can researchers efficiently develop forced aligners for regional varieties that achieve reasonably accurate alignment with a manageable amount of data? In particular, how can this be accomplished when manual annotations are scarce or entirely absent, and when text transcriptions are also lacking? To address these challenges, we propose a transferable pipeline for developing variety-specific aligners for both text-dependent and -independent alignment. 

We used Chengdu Mandarin as a case study. Chengdu Mandarin belongs linguistically to the Southwestern Mandarin group \cite{li1989mandarin} and is spoken by more than 20 million speakers. Despite having tens of millions of native speakers, Chengdu Mandarin is severely under-resourced in speech technology. The aligners we trained for this variety thus serve as valuable tools for phonetic research, and our work provides practical guidance for developing similar resources for other low-resource language varieties.

Previous studies have demonstrated the effectiveness of GMM-HMM-based systems for phonetic forced alignment~\cite{mcauliffe2017montreal}. More recently, pretrained speech encoders have been adapted for this task, although they do not consistently outperform traditional GMM-HMM approaches~\cite{Zhu2022PhoneToAudio}. In this study, we employed both approaches to develop forced aligners for Chengdu Mandarin.
Specifically, we collected approximately 17 hours of Chengdu Mandarin speech and constructed an expert-annotated grapheme-to-phoneme (G2P) dictionary tailored to its sound system. Using these resources, we first trained a GMM-HMM-based acoustic model, for \textit{text-dependent} alignment. We then applied Chengdu-MFA to the corpus to automatically generate phone-level pseudo-labels. These pseudo-labels served as supervision for fine-tuning a pretrained speech encoder on a frame-classification task, enabling \textit{text-independent} phonetic segmentation at inference time.

To assess model performance, two phoneticians manually annotated a small set of Chengdu recordings as the gold standard annotations, and compared them with the output generated by our trained models for Chengdu Mandarin as well as the baseline models pre-trained on Standard Mandarin. Alignment evaluation followed the methods in the previous studies \cite{chodroff2025comparing, mcauliffe2017montreal, Zhu2022PhoneToAudio, rousso2024tradition, rasanen2009improved, kreuk2020self}.

Our main contributions are threefold: 
(1) We release the first dedicated aligners for Chengdu Mandarin, supporting both text-dependent and text-less forced alignment, as well as a specialized G2P dictionary. 
(2) We provide empirical evidence demonstrating the limitations of applying standard-language models to regional varieties, highlighting the necessity of variety-specific training. 
(3) Most importantly, we establish and validate an end-to-end bootstrapping pipeline (G2P Dictionary \(\rightarrow\) text-dependent aligner \(\rightarrow\)pseudo-labels \(\rightarrow\) text-independent aligner) that provides the speech research community with a reproducible workflow for developing alignment tools for other under-resourced language varieties.

\section{Method}

\subsection{G2P Dictionary for Chengdu Mandarin}
While Mandarin varieties share a character-based writing system, they differ considerably in their sound inventories. The phone set used by Standard Mandarin models does not apply to the sound inventory of Chengdu Mandarin. Therefore, we compiled a Chengdu Mandarin dictionary covering all the 2876 Chinese characters in the master dataset. The dictionary was first automatically annotated using the \texttt{Pypinyin} library and DeepSeek-v3 \cite{liu2024deepseek}. Two native speakers of Chengdu Mandarin then reviewed the model-generated annotations and corrected errors.

\subsection{Chengdu-MFA}
We developed Chengdu-MFA, a GMM-HMM-based acoustic model for text-dependent forced alignment, using the Montreal Forced Aligner (MFA) framework~\cite{mcauliffe2017montreal}. In a GMM-HMM system, an utterance is represented as a sequence of hidden phonetic states, whose temporal transitions are modeled by hidden Markov models, while the acoustic distribution associated with each state is modeled using Gaussian mixture models. Given an audio recording, its orthographic transcription, and a pronunciation dictionary, the model identifies the most likely state sequence and thereby estimates word- and phone-level boundaries.

GMM-HMM systems do not require manually annotated phone boundaries for training. Instead, the model jointly estimates its acoustic parameters and latent state alignments from the utterance-level transcripts and their corresponding recordings. During forced alignment, the observed audio was constrained by the phone sequence derived from the transcript, and the most likely alignment path was decoded to obtain word- and phone-level timestamps.

In addition to performing text-dependent alignment, Chengdu-MFA served as the first stage of our bootstrapping pipeline. We applied the trained model to the Chengdu Mandarin training corpus to generate phone-level alignments automatically. 

\subsection{Chengdu-FC}
To exploit the acoustic representations learned by pretrained speech encoders for phonetic forced alignment, a prevalent strategy is to cast the alignment process as a frame-level classification problem~\cite{Zhu2022PhoneToAudio}. Given an input audio sequence of $T$ frames, the model predicts a phone label $\hat{y}_t$ for each frame $t$. The standard training objective minimizes the average frame-level cross-entropy loss:
\begin{equation}
\mathcal{L}_{\mathrm{seq}}(\mathbf{y},\hat{\mathbf{y}})
=
\frac{1}{T}
\sum_{t=1}^{T}
\mathcal{L}_{\mathrm{CE}}(y_t,\hat{y}_t),
\label{eq:standard_loss}
\end{equation}
where $\mathbf{y}=(y_1,\ldots,y_T)$ denotes the reference phone-label sequence and $\hat{\mathbf{y}}=(\hat{y}_1,\ldots,\hat{y}_T)$ denotes the corresponding model predictions. We developed Chengdu-FC model series using this learning objective with different audio encoders.

The learning objective requires speech data with frame-level phone annotations. Because manually annotated phone boundaries are costly to obtain, we used Chengdu-MFA to generate phone-level pseudo-labels for the training corpus as supervision for fine-tuning the pretrained speech encoder and its frame-classification head.

To improve the model's sensitivity to phone boundaries, we adopted a curriculum-learning strategy. The model was first trained for $E$ epochs using the standard loss in Equation~\ref{eq:standard_loss}. In subsequent epochs, frames near phone boundaries were assigned greater weights than frames in phone-internal regions. Specifically, for epochs after $E$, we used the following boundary-weighted loss:
\begin{equation}
\left\{
\begin{aligned}
\mathcal{L}_{\mathrm{seq}}(\mathbf{y},\hat{\mathbf{y}})
&=
\frac{
\sum_{t=1}^{T}
w_t\,\mathcal{L}_{\mathrm{CE}}(y_t,\hat{y}_t)
}{
\sum_{t=1}^{T} w_t
},\\
w_t
&=
1+(\gamma-1)\,
\mathbb{I}_{\mathrm{bound}}(t,r),
\end{aligned}
\right.
\label{eq:weighted_loss}
\end{equation}
where $\gamma \geq 1$ is the boundary-weighting factor and $r$ specifies the radius, measured in frames, around each reference boundary. The indicator function
$\mathbb{I}_{\mathrm{bound}}(t,r)$ equals 1 if frame $t$ lies within $r$ frames of any phone boundary and 0 otherwise. Thus, boundary-adjacent frames receive a weight of $\gamma$, whereas all other frames retain a weight of 1.

\begin{figure*}[ht]
    \centering
    \includegraphics[width=1.0\linewidth]{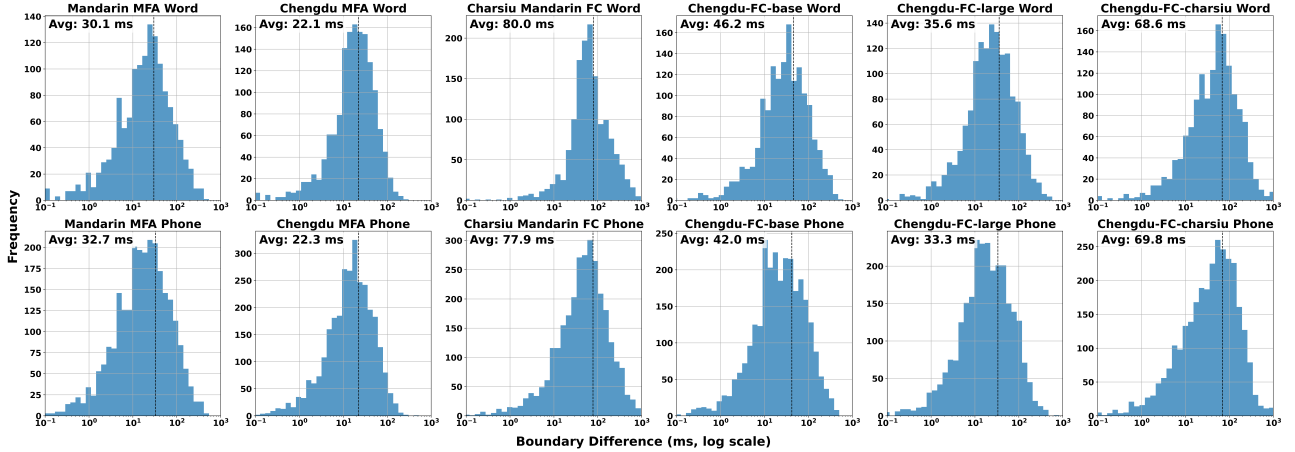}
    \captionsetup{justification=centering}
    \caption{Histograms of absolute differences (on a log scale) between forced-aligned boundaries and gold-standard annotations. The first and second rows show the word and phone tiers respectively. Dashed line represents the average boundary difference.}
    \label{fig:boundary:difference:distribution}
\end{figure*}

\section{Experiments}
\subsection{Dataset}
The master dataset for this study consisted of 15 high-quality audio recordings of Chengdu Mandarin and the corresponding text transcriptions drawn from \cite{li2022regional}. The audios were produced by 15 speakers born and raised in urban Chengdu, a southwestern city in Sichuan, China. Each audio ranges from 0.6 to 1.7 hours; the total duration is about 17.7 hours. The sampling rate of the audios is 24 kHz. 
We created subsets of this master dataset for different procedures involved in our model training and evaluation:

A \textbf{test set} was first created containing 50 minutes of recordings from the master dataset, sampled from the recordings of 10 speakers, 5 minutes per speaker. The test audios were manually aligned and transcribed at the word (Chinese characters) and phone (IPA symbols) levels by linguistic experts. 
This manual annotation serves as the gold standard against which the alignments from the baseline and our trained MFA and FC models are evaluated.

The \textbf{training set for Chengdu MFA} acoustic model took the full master dataset minus the test set. To prepare MFA input, utterance-level TextGrids were created based on text transcriptions with timestamps from ELAN \cite{Wittenburg2006ELANAP}. 
Irrelevant information such as the punctuation marks, paralinguistic annotations (e.g., ((laugh))), and other non-lexical elements were removed from the TextGrids.

We used Chengdu-MFA model to provide phonetic annotation to its training set. Of these generated phone annotations, 80$\%$\ were allocated to the \textbf{Chengdu FC training set} and 20 $\%$\ to the \textbf{Chengdu FC validation set} to select hyperparameters.

\subsection{Training Setup}
\subsubsection{Chengdu MFA}
We trained the \textbf{Chengdu-MFA} acoustic model using MFA toolkit version 3.3.3 \cite{mcauliffe2017montreal}. Following the standard practice in G2P modeling for Chinese varieties, each syllable nucleus combined with its tone was modeled as a single unit.

\subsubsection{Chengdu FC}
We fine-tined Wav2Vec2-base, Wav2Vec2-large, XLS-R-300m, Charsiu Madnarin-FC on Chengdu Mandarin respectively. We dub the finetuned models as \textbf{Chengdu-FC-base}, \textbf{Chengdu-FC-large}, \textbf{Chengdu-FC-xlsr} and \textbf{Chengdu-FC-charsiu}. 

The training was done using AdamW with a weight decay of $1\times 10^{-4}$, and a batch size of 8 on a single NVIDIA RTX 3090 GPU. The training ran for 10 epochs in total. Following a curriculum learning strategy, the first 2 epochs used the standard cross-entropy loss. For the remaining epochs, training switched to the boundary-aware weighted loss \eqref{eq:weighted_loss}.
Hyper-parameters were selected based on the classification accuracy on the validation set. We searched for the optimal learning rate within $[1\times10^{-5},3\times10^{-4}]$, boundary weight $\gamma \in[5, 15]$, and weighting radius $r\in[0,2]$. 

\subsection{Evaluation}
We performed forced alignment on the test set using the baseline Mandarin models and our trained Chengdu models. Text-dependent alignment was generated by both the MFA and FC models, while text-independent alignment was performed by FC models only. All the alignments were compared to the gold-standard manual annotations. Note that all our systems used IPA-based phone systems, therefore to ensure fair comparison across systems (our trained models vs. baselines), we treated the syllable nucleus and coda as a single unit in evaluation.

For text-dependent alignment, we measured the absolute differences between the gold standard and our alignments at the start boundaries, and presented the percentage of word and phone boundaries under different time thresholds \cite{chodroff2025comparing, mcauliffe2017montreal, rousso2024tradition}. For text-independent alignment, because boundary matching is not applicable, we evaluated the precision (p), recall (r), F1-score (F1), and R-value of the boundaries at different time tolerances \cite{Zhu2022PhoneToAudio, rasanen2009improved, kreuk2020self}.

\subsection{Result}
\subsubsection{Text-dependent alignment}
Figure~\ref{fig:boundary:difference:distribution} presents the distribution of the absolute differences between the gold standard and the aligned boundaries generated using our Chengdu Mandarin forced aligners. Both the Chengdu-MFA and -FC aligners outperformed the respective baselines at both word and phone tier. Specifically, the average boundary difference of Chengdu-MFA was 22.1 ms at the word tier, representing 26.6\% reduction compared to the Mandarin baseline. At phone tier, the average difference was 22.3 ms, a 31.8\% improvement over the baseline (Table~\ref{tab:boundary:difference:mean:median}).
Meanwhile, all Chengdu-FC models exhibited better performance than the Mandarin-FC baseline. In particular, the Chengdu-FC-xlsr achieved an average boundary difference of 32.5 ms at word tier (62.3\% shorter than the baseline) and 30.2 ms at phone tier (61.2\% shorter than the Charsiu-Mandarin-FC baseline). 

As shown in Table~\ref{tab:tolerance}, although the Mandarin-MFA model achieved the highest proportion of predictions within 10 ms, the Chengdu-MFA outperformed it at all cutoffs above 10 ms. The superior performance of Mandarin-MFA at 10 ms threshold suggests a high concentration of small errors, likely due to its large amount of training data and the pronunciation similarity of these two Mandarin varieties. 
However, it exhibited a heavier tail in the distribution, leading to a lower cumulative proportion as the tolerance threshold increases, whereas Chengdu-MFA model demonstrates more robust performance across broader tolerance ranges.

Regarding FC models in the text-dependent task, the Chengdu-FC-xlsr model demonstrated comparable performance to the MFA models. Specifically, it aligned 82.2\% of phone tier boundaries within 50 ms. This significantly outperforms the Charsiu-Mandarin-FC baseline (52.8\%) and beats the performance of the Mandarin-MFA baseline at all the time thresholds.

\begin{table}[h]
    \centering
    \captionsetup{justification=centering}
    \caption{Mean and median boundary differences between MFA and FC models (Mandarin vs. Chengdu) and the gold standard. Differences reported are all statistically significant (Welch's t-test, $p<.001$). Bold numbers mark the best in each column; underlines indicate the best FC model.}
    \begin{tabular}{lcccc}\toprule
    \multicolumn{1}{l}{Model} & \multicolumn{2}{c}{Word Tier$\downarrow$} & \multicolumn{2}{l}{Phone Tier$\downarrow$}\\
    & mean & med. & mean & med. \\ 
    \midrule
    Mandarin-MFA & 30.1 & \textbf{11.9} & 32.7 & 14.1 \\
    Chengdu-MFA (ours) & \textbf{22.1} & 13.5 & \textbf{22.3} & \textbf{12.9} \\\hdashline
    Charsiu-Mandarin-FC & 80.0 & 47.5 & 77.9 & 41.0 \\
    Chengdu-FC-charsiu (ours) & 68.6 & 34.8 & 69.8 & 35.4 \\
    Chengdu-FC-base (ours) & 46.2 & 23.3 & 42.0 & 19.6 \\
    Chengdu-FC-large (ours) & 35.6 & 17.4 & \
    33.3 & 15.4 \\
    Chengdu-FC-xlsr (ours) & \underline{32.5} & \underline{15.4} & \underline{30.2} & \underline{13.8} \\
    \midrule
    \end{tabular}
    \label{tab:boundary:difference:mean:median}
\end{table}

\begin{table}[h]
    \centering
    \captionsetup{justification=centering}
    \caption{Percentage of boundary differences within different time thresholds.}
    \begin{tabular}{lcccc}\toprule
        & \multicolumn{4}{c}{Percentage within$\uparrow$} \\
        Boundary Difference (ms) & $<10$ & $<25$ & $<50$ & $<100$ \\ 
        \midrule
        Mandarin-MFA (Word) & \textbf{.463} & .662 & .825 & .924 \\
        Mandarin-MFA (Phone) & \textbf{.414} & .643 & .814 & .924 \\
        Chengdu-MFA (Word) & .406 & \textbf{.691} & \textbf{.881} & \textbf{.978} \\
        Chengdu-MFA (Phone) & .406 & \textbf{.678} & \textbf{.868} & \textbf{.969} \\\hdashline
        Charsiu-Mandarin-FC (Word) & .258 & .327 & .528 & .776 \\
        Charsiu-Mandarin-FC (Phone) & .234 & .375 & .560 & .768 \\
        Chengdu-FC-charsiu (Word) & .293 & .439 & .597 & .793 \\
        Chengdu-FC-charsiu (Phone) & .279 & .429 & .592 & .774 \\
        
        Chengdu-FC-base (Word) & .313 & .520 & .715 & .872 \\
        Chengdu-FC-base (Phone) & .337 & .559 & .743 & .885 \\
        Chengdu-FC-large (Word) & .382 & .603 & .784 & .913 \\
        Chengdu-FC-large (Phone) & .398 & .634 & .800 & .925 \\
        Chengdu-FC-xlsr (Word) & \underline{.405} & \underline{.645}& \underline{.813} & \underline{.932} \\
        Chengdu-FC-xlsr (Phone) & \underline{.429} & \underline{.669} & \underline{.822} & \underline{.942}\\
        
        
        \midrule
    \end{tabular}
    \label{tab:tolerance}
\end{table}

\subsubsection{Text-independent alignment}
Table~\ref{tab:textless_metrics_at_tau} details the text-independent alignment performance across different time tolerances ($\tau$). Our Chengdu-FC models demonstrate a substantial advantage in the R-value. When $\tau=20$, the R-value of our Chengdu-FC-large model is 21.1\% larger than the Mandarin baseline.
While the Mandarin FC model exhibited slightly higher recall at large time tolerances (e.g., 0.954 when $\tau=100$), this comes at the cost of low precision, which indicates that the model introduced severe over-segmentation when applied to Chengdu Mandarin.

\begin{table}[h]
    \centering
    \footnotesize
    \captionsetup{justification=centering}
    \caption{Precision (p), recall (r), F1, and R-value at different time tolerances ($\tau$, ms) for Charsiu-Mandarin-FC, Chengdu-FC-base, Chengdu-FC-Charsiu, and Chengdu-FC-large models.}
    \begin{tabular}{llcccc}
        \toprule
        \textbf{$\tau$ (ms)} & \textbf{Model} & \textbf{p} & \textbf{r} & \textbf{F1} & \textbf{R-val} \\
        \midrule
        \multirow{5}{*}{20} 
            & Charsiu-Mandarin-FC & .375 & .589 & .457 & .289 \\
            & Chengdu-FC-base     & .464 & .616 & .528 & .482 \\
            & Chengdu-FC-charsiu  & .459 & .643 & .535 & .453 \\
            & Chengdu-FC-large    & \textbf{.488} & .665 & .562 & \textbf{.500} \\
            & Chengdu-FC-xlsr     & .482 & \textbf{.691} & \textbf{.567} & .463 \\
        \midrule
        \multirow{5}{*}{40} 
            & Charsiu-Mandarin-FC & .495 & .775 & .602 & .401 \\
            & Chengdu-FC-base     & .593 & .787 & .675 & .601 \\
            & Chengdu-FC-charsiu  & .588 & .824 & .685 & .566 \\
            & Chengdu-FC-large    & \textbf{.618} & .842 & \textbf{.712} & \textbf{.613} \\
            & Chengdu-FC-xlsr     & .597 & \textbf{.857} & .703 & .560 \\
        \midrule
        \multirow{5}{*}{60} 
            & Charsiu-Mandarin-FC & .563 & .882 & .684 & .455 \\
            & Chengdu-FC-base     & .641 & .849 & .728 & \textbf{.639} \\
            & Chengdu-FC-charsiu  & .629 & .882 & .733 & .597 \\
            & Chengdu-FC-large    & \textbf{.648} & .882 & \textbf{.746} & .635 \\
            & Chengdu-FC-xlsr     & .630 & \textbf{.904} & .742 & .584 \\
        \midrule
        \multirow{5}{*}{80} 
            & Charsiu-Mandarin-FC & .596 & \textbf{.932} & .725 & .478 \\
            & Chengdu-FC-base     & .663 & .880 & .754 & \textbf{.656} \\
            & Chengdu-FC-charsiu  & .649 & .910 & .756 & .611 \\
            & Chengdu-FC-large    & \textbf{.667} & .908 & \textbf{.768} & .648 \\
            & Chengdu-FC-xlsr     & .648 & .930 & .763 & .595 \\
        \midrule
        \multirow{5}{*}{100} 
            & Charsiu-Mandarin-FC & .611 & \textbf{.954} & .743 & .488 \\
            & Chengdu-FC-base     & .677 & .898 & .770 & \textbf{.666} \\
            & Chengdu-FC-charsiu  & .663 & .930 & .773 & .620 \\
            & Chengdu-FC-large    & \textbf{.681} & .927 & \textbf{.785} & .657 \\
            & Chengdu-FC-xlsr     & .660 & .947 & .777 & .603 \\
        \bottomrule
    \end{tabular}
    \label{tab:textless_metrics_at_tau}
\end{table}

Regarding training strategies, all three Chengdu-FC variants significantly outperformed the Standard Mandarin baseline (Charsiu-Mandarin-FC). Among them, Chengdu-FC-large has the largest number of parameters, followed by the Chengdu-FC-base, and then Chengdu-FC-Charsiu.
Fine-tuning from Wav2Vec2-large yielded the best performance, achieving the highest F1-scores and R-values across almost all settings. Although Chengdu-FC-Charsiu was previously fine-tuned on Mandarin data, this did not compensate for the disparity in model size, resulting in relatively reduced performance comparing to the other Chengdu-FC models.

\section{Discussion}

Our results demonstrate the benefits of variety-specific training for phonetic alignment. Under the present experimental conditions, Chengdu-MFA provided the most accurate text-dependent alignments, despite being trained on only approximately 17 hours of utterance-transcribed speech. GMM-HMM-based systems therefore remain a practical option for under-resourced language varieties when transcripts and a pronunciation dictionary are available.

Chengdu-MFA also provides an efficient means of generating phone-level pseudo-labels without manual boundary annotation. These pseudo-labels can be used to train frame-classification models for transcript-free phonetic segmentation. Although the Chengdu-FC models were generally less accurate than Chengdu-MFA in text-dependent alignment, they can operate without input transcripts at inference time. The MFA and FC approaches therefore address complementary application scenarios.

The Chengdu-specific models consistently outperformed their Standard Mandarin counterparts, confirming the benefits of variety-specific adaptation. Nevertheless, the Standard Mandarin models achieved non-trivial performance, possibly because of the phonological similarity between Standard Mandarin and Chengdu Mandarin and their large-scale training data. Cross-variety generalization also varied across modeling frameworks, with Mandarin-MFA outperforming Charsiu-Mandarin-FC in text-dependent alignment. Because the current evaluation contains speakers whose other recordings were included in training, future work should assess generalization to unseen speakers and additional speech domains.

\section{Conclusion}
In this study, we created a Chengdu Mandarin phonetic forced alignment dataset with a G2P dictionary covering all characters in the dataset. We trained a MFA acoustic model for Chengdu Mandarin for text-dependent alignment and a Chengdu Mandarin frame classification model capable of performing both text-dependent and -independent alignment. Our trained models consistently outperformed the Mandarin baselines with identical architectures but far more training data. For suggestions on applying these models, if utterance-level transcripts are available, the MFA model can provide highly accurate word and phone annotation; if not, the FC model enables effective text-independent alignment. This work contributes to the forced alignment tools for low-resource language varieties by presenting a complete training and evaluating pipeline for a specific Mandarin variety.

\newpage

\bibliographystyle{IEEEtran}
\bibliography{mybib}

\end{document}